%% file: calic_DCC.tex
 \documentclass[smallabstract,smallcaptions]{dccpaper}

\usepackage{epsfig}
\usepackage{amsmath}
\usepackage{amssymb}
\usepackage{color}
\usepackage{url}
\usepackage{graphicx}
\usepackage{amsfonts}
\usepackage{algorithmic}
\usepackage{textcomp}
\usepackage{times}
\usepackage{mathrsfs}
\usepackage{multirow}
\usepackage{multicol}
\usepackage{subfigure}

\newlength{\figurewidth}
\newlength{\smallfigurewidth}

\newcommand{\etal}{\textit{et al}.}
\newcommand{\fw}{1.9cm}
\newcommand{\fh}{3.5cm}

\begin{document}

\title
{\large
\textbf{Near-lossless $\ell_\infty$-constrained Image Decompression \\ via Deep Neural Network}
}


\author{%
Xi Zhang $^{\ast}$, Xiaolin Wu$^{\dag\ast}$ \\[0.5em]
{\small\begin{minipage}{\linewidth}\begin{center}
\begin{tabular}{ccc}
$^{\ast}$Shanghai Jiao Tong University & \hspace*{0.5in} & $^{\dag}$McMaster University \\
Shanghai, 200240, China && Hamilton, L8G 4K1, Ontario, Canada \\
\url{zhangxi_19930818@sjtu.edu.cn} && \url{xwu@ece.mcmaster.ca}
\end{tabular}
\end{center}\end{minipage}}
}

\maketitle
\thispagestyle{empty}

\begin{abstract}
Recently a number of CNN-based techniques were proposed to remove image compression artifacts. As in other restoration applications, these techniques all learn a mapping from decompressed patches to the original counterparts under the ubiquitous $\ell_2$ metric.
However, this approach is incapable of restoring distinctive image details which may be statistical outliers but have high semantic importance (e.g., tiny lesions in medical images).  To overcome this weakness, we propose to incorporate an $\ell_\infty$ fidelity criterion in the design of neural network so that no small, distinctive structures of the original image can be dropped or distorted. 
Experimental results demonstrate that the proposed method outperforms the state-of-the-art methods in $\ell_\infty$ error metric and perceptual quality, while being competitive in $\ell_2$ error metric as well.  It can restore subtle image details that are otherwise destroyed or missed by other algorithms.  Our research suggests a new machine learning paradigm of ultra high fidelity image compression that is ideally suited for applications in medicine, space, and sciences.
\end{abstract}

\section{Introduction}

In many professional applications of computer vision, such as medicine, remote sensing, sciences and precision engineering, high spatial and spectral resolutions of images are always of paramount importance.  As the achievable resolutions of modern imaging technologies steadily increase, users are inundated by the resulting astronomical amount of image data. 
For the sake of operability and cost-effectiveness, images have to be compressed for storage and communication in practical systems.

Unlike in consumer applications, such as smartphones and social media, where users are mostly interested in image esthetics, professionals of many technical fields are more concerned with the fidelity of decompressed images.  Ideally, they want mathematically lossless image compression, that is, the compression is an invertible coding scheme that can decode back to the original image, bit for bit identical.
Although the mathematically lossless image coding is the ultimate gold standard, its compression performance is too limited.  Despite years of research \cite{calic, lossless_calderbank, lossless_maniccam}, typical lossless compression ratios for medical and remote sensing images are only around 2:1, which fall far short of the requirements of most imaging and vision systems.

Due to the above difficulty, many users are forced to adopt a lossy image compression alternative, such as JPEG or JPEG 2000.  Instead of aggressive compression as in consumer applications, professional users (e.g., doctors, scientists and engineers)
set very high quality factors in lossy compression to keep compression distortions at minimum. However, strictly speaking, short of lossless coding, compression noises are inevitable; thus if and how compression noises can be removed after the decompression is a challenging and worthy research problem.  Recently, a number of machine learning-techniques are proposed to polish the decompressed images via a learnt mapping from decompressed image patches to their original counterparts \cite{ARCNN, CAR_galteri, CAR_guo}.

Apparently still being motivated by consumer applications, the above said mappings are learnt using training data under the ubiquitous $\ell_2$ metric.  This approach is incapable of restoring distinct image details lost in the compression process, which are statistical outliers but nevertheless vital to image semantics.  Such cases are common in machine vision applications; for examples, one is searching in a big ocean for a small boat, or a small lesion in a large organ. An $\ell_2$-based algorithm is prone to override such small structures by whatever dominant patterns in the background: ocean waves in the first example and liver textures in the second example.

In order to overcome the above identified drawback of existing techniques, we propose to incorporate an $\ell_\infty$ fidelity criterion in the design of a CNN for the task of compression noise removal.  The novelty is to impose a tight error bound on each single pixel so that no small structures of the original image can be dropped or distorted.  To achieve this design objective, we require the compression to be conducted in collaboration with the proposed CNN-based decompression.  Fortunately, the so-called near-lossless image compression methods \cite{near_avcibas, near_chen, near_wu}, which were developed for demanding high-fidelity applications of machine vision, suit our purpose perfectly.

The main technical development of this work is to build a deep neural network $G$ to learn a mapping from the $\ell_\infty$-encoded images to the corresponding latent images. In existing CNN-based methods for removing compression artifacts, MSE loss, perceptual loss and adversarial loss were combined, aiming at high perceptual quality.  However, for our purpose of faithfully reconstructing the latent image on a pixel-by-pixel basis, we drop the perceptual loss in the design of our anti-artifacts CNN.  More critically, we are vigilant about the side effects of the MSE loss and the adversarial loss.  The former criterion, due to the nature of $\ell_2$ metric, tends to smooth out subtle small image features; the latter criterion can fabricate false features.  These distortions, which are detrimental and should be prevented at all costs in the professional fields of medicine, space, engineering and sciences, can be suppressed by adding an $\ell_\infty$ fidelity term in the optimization of network $G$ as proposed above.  The resulting neural network $G$ is called $\ell_\infty$-CNN in the sequel.  

The above outlined $\ell_\infty$-CNN decompressor in conjunction with a collaborative $\ell_\infty$-constrained compressor presents a new paradigm of ultra high fidelity image compression.  Through extensive experiments, we demonstrate that the proposed new paradigm outperforms the state-of-the-art image compression systems in $\ell_\infty$ metric and perceptual quality for a given compression ratio, while being competitive in $\ell_2$ error metric as well.


\section{Related Works}
\label{related}
\subsection{Compression Artifacts Removal}
There is a rich body of literature on techniques for removing compression artifacts in images \cite{rw_reeve, rw_zhai, rw_alter, rw_liu}. The majority of the studies on the subject focus on postprocessing JPEG images to alleviate compression noises, apparently because JPEG is the most widely used lossy compression method.

Reeve \etal \cite{rw_reeve} proposed to remove structured discontinuities of DCT code blocks by Gaussian filtering of the pixels around the DCT block boundaries.  This work was improved by
Zhai \etal \cite{rw_zhai} who performed postfiltering in shifted overlapped windows and fused the filtering results.  
A total minimum variation method constrained by the JPEG quantization intervals was used by Alter \etal \cite{rw_alter} to reduce blocking artifacts and Gibbs phenomenon while preserving sharp edges.

Given the recent rapid development of deep convolutional neural networks (CNN), a number of CNN-based compression artifacts removal methods were published.
Borrowing the CNN for super-resolution (SRCNN), Dong \etal \cite{ARCNN} proposed an artifact reduction CNN (ARCNN). 
It was improved by Svoboda \etal \cite{rw_svoboda} who combined residual learning and symmetric weight initialization.
Recently, Guo \etal \cite{CAR_guo} and Galteri \etal \cite{CAR_galteri} proposed to reduce compression artifacts by Generative Adversarial Network (GAN), as GAN is able to generate sharper image details. 
It should be noted, however, that the GAN results may fabricate a lot of false hallucinated details, which is strictly forbidden in many scientific and medical applications.

\subsection{Near-lossless Image Coding}

In the compression literature, near-lossless image coding refers to the $\ell_\infty$-constrained compression schemes that guarantee the compression error to be no larger than a user-specified bound for every pixel.  This can be realized within the framework of classic predictive coding. 
Denoting by $X$ an image and $x_i$ the value of pixel $i$, image $X$ is compressed pixel by pixel sequentially, by first making a prediction of $x_i$:
\begin{align}
\tilde{x}_i = F(C_i)
\end{align}
where $C_i$ is a causal context that consists of previously coded pixels adjacent to $x_i$, and then entropy encoding and transmitting the prediction residual
$e_i = x_i - \tilde{x}_i$.
To gain higher compression ratio, one can quantize $e_i$ uniformly in step size $\tau$ to $\hat{e}_i$.
In this way, the decoded pixel value becomes $y_i = \hat{e}_i + \tilde{x}_i$, with quantization error
\begin{align}
d = x_i - y_i = e_i - \hat{e}_i
\end{align}
Due to the quantization step size $\tau$, the quantization error will be no greater than the bound $\tau$ for every pixel:
\begin{align}
-\tau \leq x_i - y_i \leq \tau
\label{bound}
\end{align}
The above inequalities not only impose an $\ell_\infty$ error bound, but more importantly they, for the purpose of this work, provide highly effective priors, on per pixel, to optimize the deep neural networks for compression artifacts removal.

\section{Methodology}
\label{methodology}

\subsection{Problem Formulation}
Inheriting the symbols introduced above, $X$ denotes the original image, $Y = A^{-1}A(X)$ the decompression result of compressing $X$ by a near-lossless compression algorithm $A$.
In the restoration of decompressed image $Y$, the aim is to compute a refined reconstruction image $\hat{X}$ from a decompressed image $Y$ by maximally removing compression artifacts in $Y$.

To solve the problem of compression artifacts removal, we train a neural network $\ell_\infty$-CNN (denoted by $G$) that takes decompressed image $Y$ as its input and returns the restored image $\hat{X} = G(Y)$.  In order to satisfy the stringent fidelity requirements of medical and scientific applications, the final output image
$\hat{X}$ needs to be close not only perceptually but also mathematically
to the original image $X$.  Having this design objective in mind, we optimize the $\ell_\infty$-CNN with a new cost function $L_G(X, G(Y))$:
\begin{align}
G = \arg \min_{G} \sum^{N}_{n=1} L_G(X_n, G(Y_n)),
\end{align}
for a given training set containing $N$ samples $\{(X_n; Y_n)\}_{1 \leq n \leq N}$.  Compared with existing works, the new cost function $L_G$, to be discussed in detail in the next section, adds an $\ell_\infty$ error bound in the reconstruction process.


\subsection{Network Architecture}

Considering that an original image and its decompressed version populate on different manifolds, we propose to use a generative adversarial network (GAN) \cite{GAN, DCGAN} to perform the task of compression artifacts removal.
For our task, one of image restoration, network $\ell_\infty$-CNN outlined above, as the generator, is tuned to pass the test of a discriminative network $D$.  The latter network is trained to distinguish the output results of the former network from original images.

\begin{figure}[!t]
\centering
	\subfigure[The generative network $\ell_\infty$-CNN.]{
		\includegraphics[width=6cm, height=10cm]{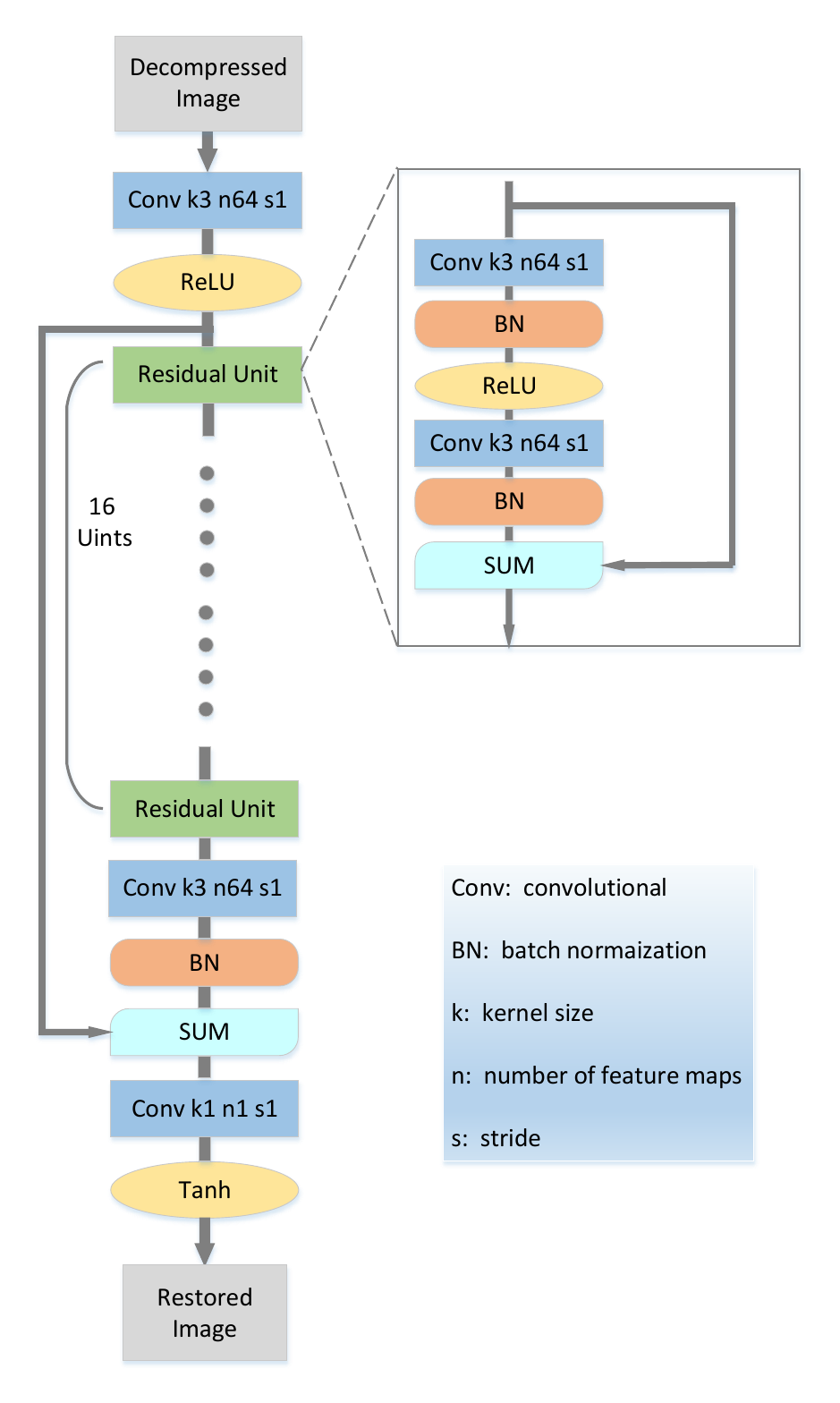}
		\label{netG}
	}
	\hskip 0.5cm
	\subfigure[The discriminative network $D$.]{
		\includegraphics[width=6cm, height=10cm]{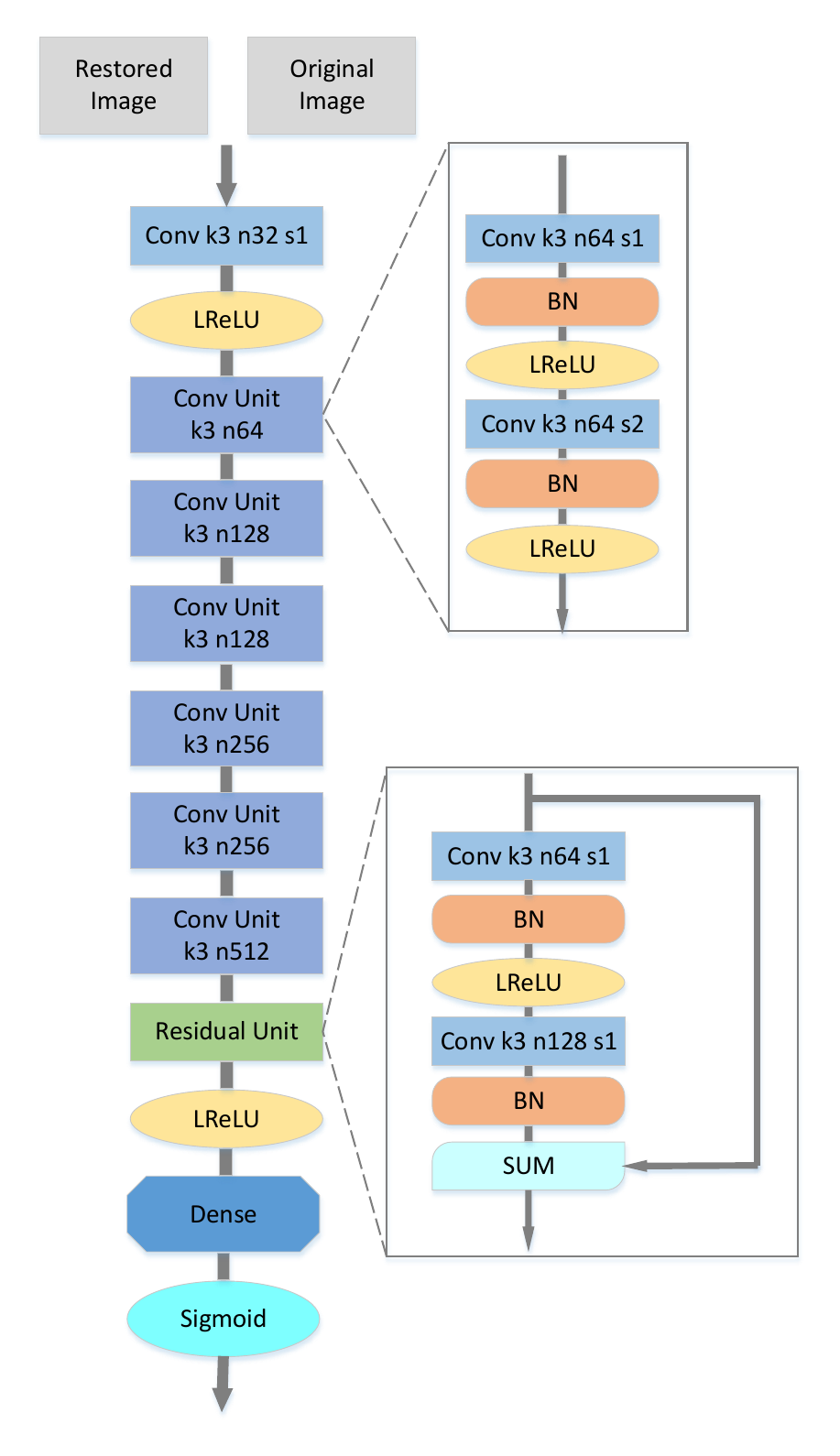}
	\label{netD}
	}
\vskip -0.5cm
\caption{The illustration of the generative network $\ell_\infty$-CNN and the discriminative network $D$.}
\vskip -0.5cm
\end{figure}

Recent works show that deeper neural network architectures \cite{VGG, GoogleNet} can achieve superior results, as they have sufficient capacity to learn a mapping with very high complexity.
In addition, networks designed with residual units containing skip connections \cite{ResNet} have achieved the state-of-the-art results for many high-level vision tasks like recognition, segmentation and low-level vision tasks like super-resolution, etc.
Built upon the past successes of others, we adopt the residual units in the construction of our deep neural network
$\ell_\infty$-CNN for near-lossless $\ell_\infty$-constrained image decompression.
Like in SRGAN \cite{SRGAN}, our generative network $\ell_\infty$-CNN contains 16 residual units. Each residual unit consists of two convolutional layers with small $3 \times 3$ kernels and 64 feature maps, followed by batch-normalization layers and ReLU activation layers. The integral architecture of network $\ell_\infty$-CNN is illustrated in Figure \ref{netG}. The discriminative network $D$ will be introduced in the next section.

\section{Loss Function}
\label{loss}
In this section, we will discuss the proposed loss function $L_G$ in detail.
Generally, $L_G$ consists of three parts: MSE loss, adversarial loss and $\ell_\infty$-constrained loss.
The three loss functions are adopted to jointly optimize the proposed $\ell_\infty$-CNN.

\subsection{MSE loss}
Although Mean Squared Error (MSE) loss $L_{mse}$ is the most widely used loss function in image restoration algorithms, it has been criticized for producing overly smooth results.
Solely minimizing MSE seeks a good approximation in average sense, but it takes the risk of destroying distinctive image details which may be statistical outliers but have high semantic importance.


\subsection{Adversarial loss}
Following Goodfellow \etal's wisdom, we define a discriminative network $D$ optimized along with the proposed $\ell_\infty$-CNN to solve the following min-max problem:
\begin{align}
\min_G \max_D V(G,D) =  \mathbb{E}_{X} \big[ log D(X) \big] + \mathbb{E}_{Y} \big[ log(1-D(G(Y))) \big]
\end{align}
In practice, for better gradient behavior, we minimize $-log D(G(Y))$ instead of $log(1-D(G(Y)))$, as proposed in \cite{GAN}.
Therefore, the adversarial loss for training generative network $\ell_\infty$-CNN is defined as:
\begin{align}
L_{adv} = -log D(G(Y))
\end{align}
and the loss for optimizing discriminative network $D$ is defined as:
\begin{align}
L_D = - \big[ log(D(X))+log(1-D(G(Y))) \big]
\label{loss_D}
\end{align}

In our task, minimizing $L_{adv}$ forces the generative network $\ell_\infty$-CNN to produce restored images that cannot be distinguished from the original images by $D$.
Meanwhile, minimizing $L_D$ forces the discriminative network $D$ to distinguish whether an image is original image or the restored image produced by $\ell_\infty$-CNN.

The architecture of our discriminative network $D$ is improved on the discriminator of SRGAN \cite{SRGAN}.
Specifically, we deepen the network and add a residual unit before the dense layer.
As shown in Figure \ref{netD}, the discriminator $D$ contains six convolutional units and a residual unit. Each convolutional unit consists of two convolutional layers, respectively followed by a batch-normalization layer and a LeakyReLU activation layer. The newly added residual unit contains two convolutional layers, and each of them is followed by a batch-normalization layer. For LeakyReLU, the slope of the leak is set to $0.2$.



\subsection{$\ell_{\infty}$-constrained loss}
\label{infty_loss}
To overcome the above identified weaknesses of MSE loss and adversarial training, we incorporate the $\ell_\infty$ fidelity criterion of near-lossless compression into the optimization of $\ell_\infty$-CNN.
The strict error bound restriction will force $\ell_\infty$-CNN to produce reconstructed results with restored small structures and less hallucinated details.

Given the error bound $\tau$ of near-lossless compression algorithm and the decompressed image $Y$, the reconstructed image $\hat{X}$ should satisfy the following constraint:
\begin{align}
y_i - \tau \leq \hat{x}_i \leq y_i + \tau
\label{x_i}
\end{align}
where $i$ traverse all pixels in $\hat{X}$ and $Y$.


For pixels in $\hat{X}$, the $\ell_\infty$-constrained loss penalizes the pixel values that are out of the range $[y_i-\tau, y_i+\tau]$, but does not affect those that are in the range $[y_i-\tau, y_i+\tau]$.

For this goal, we rewrite the Eq(\ref{x_i}) as:
\begin{align}
| \hat{x}_i - y_i | \leq \tau
\end{align}
and formulate the elaborate $\ell_\infty$-constrained loss as following:
\begin{align}
L_\infty = -\frac{1}{WH} \sum_i log \big[ 1 -  max \big( |\hat{x}_i - y_i| - \tau, 0 \big) \big]
\end{align}

The $\ell_\infty$-constrained loss will increase rapidly when pixel values are out of range $[y_i-\tau, y_i+\tau]$. This severe penalty can force $\ell_\infty$-CNN to produce results satisfying $\ell_\infty$ constraint.


\subsection{Joint optimization}
We combine the aforementioned three loss functions to jointly optimize the proposed $\ell_\infty$-CNN.
The joint loss function is defined as:
\begin{align}
  L_{G} = L_{mse} + \lambda_1 L_{adv} + \lambda_2 L_{\infty}
\end{align}
where $\lambda_1$ and $\lambda_2$ are hyper-parameters.


\section{Training of $\ell_\infty$-CNN}
\label{training}
In this section, we present the details of training the proposed  $\ell_\infty$-CNN for near-lossless image decompression. 




\subsection{Dataset}
In the existing works on CNN-based compression artifacts removal \cite{CAR_guo,CAR_galteri,ARCNN}, data used for training are from the popular datasets like BSD100, ImageNet or MSCOCO.
However, because images in these datasets are already compressed and have relatively low resolutions, they are not suitable as the ground truth for our purpose of ultra high fidelity image compression for professional applications. Instead we choose the high-quality image dataset DIV2K for training. The DIV2K dataset is a newly proposed 2K resolution image dataset for image restoration tasks.
In the collaborative compression phase, we adopt the $\ell_\infty$-constrained (or near-lossless) CALIC \cite{near_wu} to guarantee the compression error to be no larger than a specified bound $\tau$ for every pixel ($\tau=6, 8, 10$ used in our experiments).


\subsection{Training Details}
Images in the DIV2K dataset are decomposed into $64 \times 64$ sub-images with stride $32$, after compressed by the near-lossless CALIC algorithm.
In the training phase, we train our networks using Adam optimizer with momentum term $\beta_1=0.9$.
The neural networks are trained with $100$ epochs at the learning rate of $10^{-4}$ and other $50$ epochs with learning rate of $10^{-5}$.
To ensure the stability of the adversarial training, we perform the one-sided label smoothing for the discriminative network $D$ proposed in \cite{GAN_smooth}.
It is noteworthy that we initialize the proposed $\ell_\infty$-CNN without adversarial loss in the first $50$ epochs. The reason is, at the early stage of training, images reconstructed by $\ell_\infty$-CNN can be distinguished easily by discriminator $D$.
It is unnecessary to feed the reconstructed images to discriminator $D$ at the early stage.

\section{Performance Evaluation}
\label{performance}
We have implemented the proposed $\ell_\infty$-CNN for the task of removing compression artifacts, and conducted extensive experiments of near-lossless image decompression with it.  In this section, we compare our results with two popular image compression algorithms: JPEG 2000 \cite{jpeg2000} and WebP \cite{webp}.
Webp is an image format developed by Google, announced in 2010 as a new open standard for image compression.
To compare fairly, for each test image, the rates of JPEG2000 and WebP are adjusted to match that of the near-lossless CALIC.

In addition, we also compare the new $\ell_\infty$-CNN with ARCNN after training the latter with our dataset.  Unfortunately, we cannot compare with other CNN-based methods for compression artifacts removal because of no access to the authors' source codes.
In order to isolate the effects of imposing the $\ell_\infty$ error bounds,
we also train a baseline using only the MSE loss under our network architecture, and compare the baseline with the proposed $\ell_\infty$-CNN.


The commonly used high-quality image dataset LIVE1 is used to evaluate the above mentioned methods.  In addition, we add a set of high-resolution aerial and satellite images as test images, for they are typical of those in professional applications.
Performance results of the competing methods are tabulated in Tables \ref{tab:aerial},\ref{tab:live1}. 
As demonstrated in Tables \ref{tab:aerial}, \ref{tab:live1}, the proposed $\ell_\infty$-CNN outperforms JPEG2000, WebP and ARCNN consistently. 

As expected, the baseline model trained using only the MSE loss achieves the highest PSNR.  Because maximizing PSNR is equivalent to minimizing the MSE loss, when measured by PSNR, a model trained to minimize the MSE error should always outperform a model that minimizes any other losses.
However, remarkably, after adding the $\ell_\infty$ fidelity criterion, the $\ell_\infty$-CNN achieves a much tighter $\ell_\infty$ bound and the better performances in the SSIM measurement, without materially affecting the $\ell_2$ error.

\input{tabs/tab_aerial}

\input{tabs/tab_live1}

\begin{figure*}[!h]
\centering
\subfigure[Image]{\includegraphics[height=3.5cm,width=3cm]{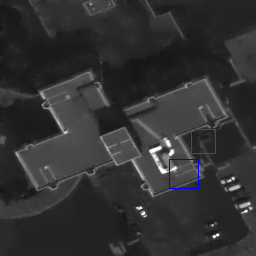}}
\subfigure[Original]{\includegraphics[height=\fh,width=\fw]{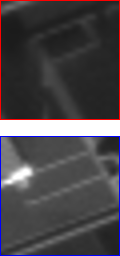}}
\subfigure[CALIC]{\includegraphics[height=\fh,width=\fw]{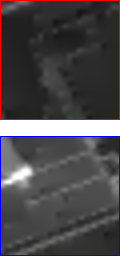}}
\subfigure[JPG2000]{\includegraphics[height=\fh,width=\fw]{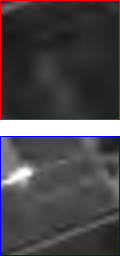}} 
\subfigure[WebP]{\includegraphics[height=\fh,width=\fw]{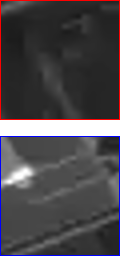}}
\subfigure[ARCNN]{\includegraphics[height=\fh,width=\fw]{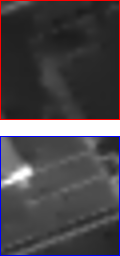}}
\subfigure[$\ell_\infty$-CNN]{\includegraphics[height=\fh,width=\fw]{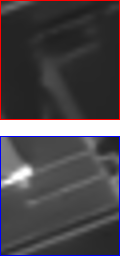}}
\vskip -0.6cm
\caption{Comparisons with the state-of-the-art methods on an aerial image.}
\label{demo1}
\end{figure*}
\begin{figure*}[!h]
\centering
\subfigure[Image]{\includegraphics[height=3.5cm,width=3cm]{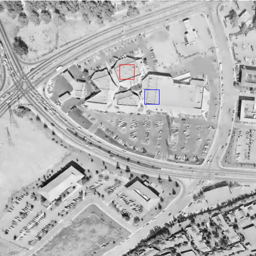}}
\subfigure[Original]{\includegraphics[height=\fh,width=\fw]{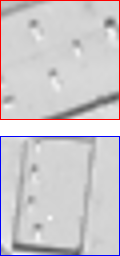}}
\subfigure[CALIC]{\includegraphics[height=\fh,width=\fw]{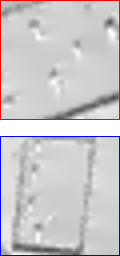}}
\subfigure[JPG2000]{\includegraphics[height=\fh,width=\fw]{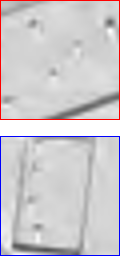}} 
\subfigure[WebP]{\includegraphics[height=\fh,width=\fw]{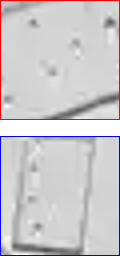}}
\subfigure[ARCNN]{\includegraphics[height=\fh,width=\fw]{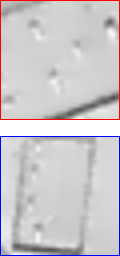}}
\subfigure[$\ell_\infty$-CNN]{\includegraphics[height=\fh,width=\fw]{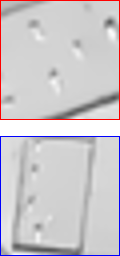}}
\vskip -0.6cm
\caption{Comparisons with the state-of-the-art methods on an aerial image.}
\label{demo2}
\end{figure*}

Finally, we would like to bring readers' attention to the enlarged parts of Figure~\ref{demo1},~\ref{demo2}.
In the red window of Figure~\ref{demo1}, the original image has a rectangular structure that is severely degraded by the near-lossless CALIC, and completely removed by JPEG 2000 and WebP.  ARCNN fails to repair the degradation of CALIC and only makes the image more blurred.  In contrast, the proposed $\ell_\infty$-CNN is able to recover the original structure from the poor near-lossless CALIC decompressed image.
In the blue window of Figure~\ref{demo1}, the lines are jagged in the near-lossless CALIC and WebP images, and almost erased in the JPEG 2000 image.  Here ARCNN fails to remove the jaggy artifacts and further blurs the lines. Only the $\ell_\infty$-CNN is able to recover the lines faithfully.

In the red and blue windows of Figure~\ref{demo2}, the sharp dent structures are visibly distorted by near-lossless CALIC, and they are made smaller and smoother by JPEG 2000 and WebP; these compression distortions can lead to misjudgments of the size, depth, and shape of these structures on the ground.  While ARCNN does not cure these problems but again blurs the image, the proposed image recovers the ground truth flawlessly.

\Section{References}
\bibliographystyle{IEEEbib}
\bibliography{calic_ref}

\end{document}

%% file: tabs/tab_aerial.tex
\begin{table*}[!t]
\caption{Comparisons with the state-of-the-art methods on the aerial image set.}
\begin{center}
\renewcommand{\multirowsetup}{\centering}
\begin{footnotesize}
\begin{tabular}{ccccccccc}
\hline
$\tau$ & Measure & CALIC & JPEG2000 & WebP & ARCNN & MSE & $\ell_\infty$-CNN \\ 
\hline
\hline
         & PSNR & $37.35$ & $38.35$ & $38.23$ & $38.37$ & $\bf{39.60}$ & $39.52$ \\
$\tau$=6 & SSIM & $0.9411$ & $0.9487$ & $0.9538$ & $0.9516$ & $0.9620$ & $\bf{0.9631}$ \\
         & $\ell_\infty$ bound & $6$ & $18.32$ & $18.34$ & $11.24$ & $11.25$ & $\bf{10.95}$ \\
\hline
         & PSNR & $35.15$ & $36.91$ & $36.72$ & $36.96$ & $\bf{37.84}$ & $37.76$ \\
$\tau$=8 & SSIM & $0.9131$ & $0.9327$ & $0.9376$ & $0.9378$ & $0.9462$ & $\bf{0.9475}$ \\
         & $\ell_\infty$ bound & $8$ & $25.05$ & $23.59$ & $15.05$ & $14.89$ & $\bf{14.24}$ \\
\hline
          & PSNR & $33.49$ & $35.68$ & $35.56$ & $35.61$ & $\bf{36.39}$ & $36.32$ \\
$\tau$=10 & SSIM & $0.8880$ & $0.9159$ & $0.9213$ & $0.9199$ & $0.9293$ & $\bf{0.9304}$ \\
          & $\ell_\infty$ bound & $10$ & $28.09$ & $27.14$ & $18.36$ & $18.05$ & $\bf{17.87}$ \\
\hline
\end{tabular}
\end{footnotesize}
\end{center}
\label{tab:aerial}
\vskip -0.5cm
\end{table*}

%% file: tabs/tab_live1.tex
\begin{table*}[!t]
\caption{Comparisons with the state-of-the-art methods on dataset LIVE1.}
\begin{center}
\renewcommand{\multirowsetup}{\centering}
\begin{footnotesize}
\begin{tabular}{cccccccccc}
\hline
$\tau$ & Measure & CALIC & JPEG2000 & WebP & ARCNN & MSE & $\ell_\infty$-CNN \\ 
\hline
\hline
         & PSNR & $37.13$ & $38.32$ & $38.11$ & $38.05$ & $\bf{39.61}$ & $39.49$ \\
$\tau$=6 & SSIM & $0.9503$ & $0.9623$ & $0.9668$ & $0.9627$ & $0.9731$ & $\bf{0.9745}$ \\         
         & $\ell_\infty$ Bound & $6$ & $20.66$ & $19.10$ & $11.83$ & $11.79$ & $\bf{11.54}$ \\
\hline
         & PSNR & $35.05$ & $36.62$ & $36.55$ & $36.65$ & $\bf{37.84}$ & $37.72$ \\
$\tau$=8 & SSIM & $0.9299$ & $0.9496$ & $0.9563$ & $0.9546$ & $0.9632$ & $\bf{0.9643}$ \\
         & $\ell_\infty$ bound & $8$ & $26.34$ & $24.28$ & $15.86$ & $15.37$ & $\bf{14.34}$ \\
\hline
          & PSNR & $33.21$ & $35.23$ & $35.21$ & $35.02$ & $\bf{36.38}$ & $36.24$ \\
$\tau$=10 & SSIM & $0.9108$ & $0.9359$ & $0.9449$ & $0.9407$ & $0.9518$ & $\bf{0.9532}$ \\
          & $\ell_\infty$ bound & $10$ & $31.41$ & $29.17$ & $19.34$ & $18.61$ & $\bf{18.21}$ \\
\hline
\end{tabular}
\end{footnotesize}
\end{center}
\label{tab:live1}
\vskip -0.5cm
\end{table*}